\documentclass[letterpaper, 10 pt, conference]{ieeeconf} 
\IEEEoverridecommandlockouts
\usepackage{cite}
\usepackage{comment}
\usepackage{amsmath,amssymb,amsfonts}
\usepackage{tabularx} 
\usepackage{algorithmic}
\usepackage{graphicx}
\usepackage{textcomp}
\usepackage{xcolor}
\usepackage{multirow}
\usepackage{etoolbox}
\usepackage{balance}
\usepackage[nolist,nohyperlinks]{acronym}
\usepackage{tikz}
\usepackage{dsfont}
\usepackage{url}            
\usepackage{booktabs}       
\usepackage{nicefrac}       
\usepackage{microtype}      
\usepackage{xcolor}         
\makeatletter
\patchcmd{\@makecaption}
  {\scshape}
  {}
  {}
  {}
\makeatother

\def\BibTeX{{\rm B\kern-.05em{\sc i\kern-.025em b}\kern-.08em
    T\kern-.1667em\lower.7ex\hbox{E}\kern-.125emX}}
\usetikzlibrary{
  shapes,arrows,automata,fit,backgrounds,calc,positioning,patterns,
  decorations.pathreplacing,
  decorations.pathmorphing
}

\begin{acronym}
\acro{OC}{optimal control}
\acro{LQR}{linear quadratic regulator}
\acro{MAV}{micro aerial vehicle}
\acro{IMAV}{insect-scale micro aerial vehicle}
\acro{GPS}{guided policy search}
\acro{UAV}{unmanned aerial vehicle}
\acro{MPC}{model predictive control}
\acro{BC}{behavior cloning}
\acro{DR}{domain randomization}
\acro{IL}{imitation learning}
\acro{DAgger}{dataset aggregation}
\acro{MDP}{Markov decision process}
\acro{CoM}{Center of Mass}
\acro{SD}{standard deviation}
\acro{MSE}{mean squared error}
\acro{RMSE}{root mean square error}
\acro{NN}{neural network}
\acro{UKF}{unscented Kalman filter}
\acro{PPO}{proximal policy optimization}
\acro{LfD}{learning from demonstration}
\acro{RL}{reinforcement learning}
\acro{PPO}{proximal policy optimization}
\acro{DEA}{dielectric elastomer actuator}
\acro{LSTM}{long short-term memory}
\acro{MoI}{moment of inertia}
\end{acronym}

\def\-{\raisebox{.75pt}{-}}
    
\begin{document}

\setlength{\abovedisplayskip}{3pt}
\setlength{\belowdisplayskip}{3pt}


\title{\vspace*{-0mm} \LARGE \bf Hovering Flight of Soft-Actuated Insect-Scale Micro Aerial Vehicles using Deep Reinforcement Learning \vspace*{-0mm}

\author{Yi-Hsuan Hsiao$^\star$, Wei-Tung Chen$^\star$, Yun-Sheng Chang$^\star$, Pulkit Agrawal, and YuFeng Chen$^\dagger$}

\thanks{$^\star$These authors contributed to this work equally. $^\dagger$Corresponding author.}

\thanks{These authors are with Massachusetts Institute of Technology (MIT), Cambridge, MA, USA (email: \texttt{yhhsiao, weitung, yschang, pulkitag, yufengc@mit.edu}).}

\thanks{This work was partially supported by the National Science Foundation (FRR-2202477 and FRR-2236708), and the Research Laboratory of Electronics, MIT (2244181). Any opinions, findings, and conclusions or recommendations expressed in this material are those of the authors and do not necessarily reflect the views of the National Science Foundation. Yi-Hsuan Hsiao is supported by the Mathworks Engineering Fellowship.}

}

\maketitle

\begin{abstract}

Soft-actuated insect-scale micro aerial vehicles (IMAVs) pose unique challenges for designing robust and computationally efficient controllers. At the millimeter scale, fast robot dynamics ($\sim$ms), together with system delay, model uncertainty, and external disturbances significantly affect flight performances. Here, we design a deep reinforcement learning (RL) controller that addresses system delay and uncertainties. To initialize this neural network (NN) controller, we propose a modified behavior cloning (BC) approach with state-action re-matching to account for delay and domain-randomized expert demonstration to tackle uncertainty. Then we apply proximal policy optimization (PPO) to fine-tune the policy during RL, enhancing performance and smoothing commands. In simulations, our modified BC substantially increases the mean reward compared to baseline BC; and RL with PPO improves flight quality and reduces command fluctuations. We deploy this controller on two different insect-scale aerial robots that weigh 720 mg and 850 mg, respectively. The robots demonstrate multiple successful zero-shot hovering flights, with the longest lasting 50 seconds and root-mean-square errors of 1.34 cm in lateral direction and 0.05 cm in altitude, marking the first end-to-end deep RL-based flight on soft-driven IMAVs.

\end{abstract}



\section{Introduction}
Inspired by the exquisite maneuverability of natural insects, the robotics community has developed \acp{IMAV} that weigh less than a gram and are capable of stable hovering \cite{ma2013controlled,chukewad2021robofly,bena2023high,chen2019controlled}. Among these platforms, a class of soft-actuated \acp{IMAV} \cite{chen2019controlled} has gained particular attention due to their resilience to collisions \cite{chen2021collision}. Driven by muscle-like \acp{DEA}, these soft-actuated sub-gram robots can absorb external impacts, highlighting the potential of IMAV applications such as assisted pollination in unstructured environments. 

Despite demonstrating unique capabilities, these DEA-driven \acp{IMAV} face distinct challenges in flight controller design due to their soft actuation. First, while the soft robots exhibit excellent impact resistance, they respond more slowly and require real-time communication between the off-board sensing, power, and control subsystems. Prior work \cite{chen2019controlled} reported a 15- to 20-ms system delay that is contributed by the soft actuation and the communication between the robot and external apparatus. Such delay is critical to \acp{IMAV}, which have fast body dynamics in the millisecond range. Traditional model-based flight controllers\cite{chen2019controlled}\cite{chen2021collision} mitigate this issue by setting non-aggressive control gains; however, this method greatly reduces the closed-loop control performance, making it difficult to track aggressive and long trajectories, like the ones shown on larger scale flying robots \cite{song2023reaching}.

The second control challenge comes from model uncertainty and large external disturbances. The fabrication of soft IMAVs requires manual assembly,
which leads to a 10-20$\%$ error in the estimation of the robot's \ac{MoI}. 
In addition, the soft actuators require 1.4- to 2-kV voltage for flights, so these DEA-driven IMAVs are tethered to a bundle of wires for offboard power during their aerial maneuvers. These wires contribute to position and attitude-dependent disturbances that are difficult to model.


\begin{figure}[t]
\centering
\centerline{\includegraphics[width=0.485\textwidth]{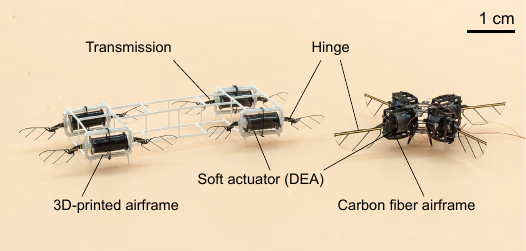}}
\vspace{-2.5mm}
\caption{An image of a 720-mg eight-wing micro-aerial-robot (left) and an 850-mg four-wing micro-aerial-robot (right) both driven by DEAs. The robot consists of either a 3D-printed or a carbon fiber airframe that connects four modules. Each module has a DEA, transmissions, wing hinges, and wings. The robot requires external systems for sensing, control, and power.  }
\label{fig:robot_photo}
\vspace{-6mm}
\end{figure}


\begin{figure*}[t]
    \centering
    \scalebox{0.95}{\begin{tikzpicture}

\definecolor{red}{RGB}{181, 23, 0}
\definecolor{blue}{RGB}{0, 118, 186}
\definecolor{gray}{RGB}{146, 146, 146}
\definecolor{green}{RGB}{56, 142, 60}

\tikzset{
  pics/pics-multi-bg/.style n args={1}{
    code={
      \node[
        inner sep=0,
        fill=black!2,
      ] (#1) {};

      \node[
        fit=(#1),
        inner sep=0,
        pattern=north east lines, pattern color=black!15,
      ] {};
    }
  }
}

\tikzset{
  style-pi/.style={
    draw,
    fill=yellow!10,
    rounded corners,
    minimum width=10mm,
    minimum height=8mm,
  },
  style-fit/.style={
    draw,
    dashed,
    inner sep=3mm,
  },
  style-traj-elem/.style={
    minimum width=0,
    inner sep=0
  },
  style-gentraj/.style={
    ->,
    decorate, decoration={
      snake,
      segment length=3mm,
      amplitude=1pt,
      post length=1pt,
    },
  }
}

\def\spacing{4mm}

\footnotesize

%
%
\node[style-pi] (pi-expert-0) {$\pi_{\color{red}e_0}$};
\node[below=3mm of pi-expert-0, style-pi] (pi-expert-1) {$\pi_{\color{red}e_1}$};
\node[below=6mm of pi-expert-1, style-pi] (pi-expert-n) {$\pi_{\color{red}e_n}$};
\node[] at ($(pi-expert-1)!0.5!(pi-expert-n)$) {\rotatebox{90}{$\cdots$}};
\node[fit=(pi-expert-0)(pi-expert-n)] (bg-pi-expert) {};

%
%
\draw pic[
  right=10mm of bg-pi-expert,
  minimum width=19mm,
  minimum height=24mm,
] {pics-multi-bg={bg-trajs}};

\node[
  below=4mm of bg-trajs.north west,
  anchor=north west,
  matrix,
  minimum width=19mm,
  draw,
  fill=black!2,
] {
  \scriptsize
  \node[style-traj-elem] (s0-traj) {$\mathbf{s}_0$,};
  \node[right=0mm of s0-traj, style-traj-elem] (a0-traj) {$\mathbf{a}_0$,};
  \node[right=0mm of a0-traj, style-traj-elem] (s1-traj) {$\mathbf{s}_1$};
  \node[right=-1pt of s1-traj, style-traj-elem] (dots-traj) {...};
  \node[right=0mm of dots-traj, style-traj-elem] (ad-traj) {$\mathbf{a}_d$};
  \node[right=-1pt of ad-traj, style-traj-elem] {...};
  \\
};

%
%
\draw[style-gentraj] (pi-expert-0.east) -- ($(pi-expert-0.east -| bg-trajs.west) - (0, 5mm)$);
\draw[style-gentraj] (pi-expert-1.east) -- ($(pi-expert-1.east -| bg-trajs.west) - (0, 5mm)$);
\draw[style-gentraj] (pi-expert-n.east) -- ($(pi-expert-n.east -| bg-trajs.west) + (0, 5mm)$);

%
%
\draw pic[
  right=\spacing of bg-trajs,
  minimum width=15mm,
  minimum height=36mm,
] {pics-multi-bg={bg-pairs}};
\node[
  below=0mm of bg-pairs.north west,
  anchor=north west,
  fill=black!2,
  draw,
  minimum width=15mm,
  inner sep=2pt,
] (pair-0) {\scriptsize $(\mathbf{s}_{\color{red}0}, \mathbf{a}_{\color{red}d})$};
\node[
  below=5mm of pair-0,
  fill=black!2,
  draw,
  minimum width=15mm,
  inner sep=2pt,
] (pair-1) {\scriptsize $(\mathbf{s}_{\color{red}t}, \mathbf{a}_{\color{red}t + d})$};

%
%
\draw[->] (s0-traj.north) to[out=60,in=190] (pair-0.west);
\draw[->] (ad-traj.north) to[out=80,in=190] (pair-0.west);

%
%
\node[
  right=11mm of bg-pairs,
  style-pi
] (pi-bc) {$\pi_{\theta}$};

%
%
\draw[->] (bg-pairs) -- node[fill=blue!7] {BC} (pi-bc);

%
%
\node[
  right=11mm of pi-bc,
  style-pi
] (pi-ppo) {$\pi_{\theta'}$};

%
%
\draw[->] (pi-bc) -- node[fill=blue!7] (ppo) {PPO} (pi-ppo);
\node[
  below=8mm of ppo,
  inner sep=0,
] (reward) {$r_{\color{red}k_p, k_e, \dotsc}$};
\draw[->] (reward) -- (ppo);

%
%
\tikzset{
  style-realworld/.style={
    draw,
    fill=green!4,
    rounded corners,
  },
}

\node[
  right=\spacing of pi-ppo,
  style-realworld,
  minimum width=15mm,
  minimum height=18mm,
] (simulink) {};
\node[
  anchor=north west,
] at (simulink.north west) {Simulink};
\node[
  above left=1mm and 1mm of simulink.south east,
  anchor=south east,
  draw,
  dotted,
  rounded corners,
  fill=yellow!10,
  minimum width=7mm,
  minimum height=5.6mm,
] (simulink-pi) {};

\node[
  right=4mm of simulink,
  style-realworld,
  minimum width=15mm,
  minimum height=18mm,
] (sensor) {};
\node[
  anchor=north west,
] at (sensor.north west) {Vicon};

\node[
  minimum width=8mm,
  minimum height=5mm,
  draw,
  rounded corners,
  fill=green!10,
] at (sensor.center) (robot) {Robot};

\draw[->] (simulink) -- node[above, opacity=0] (anchor-a-y) {$a$} (robot);
\draw[->, rounded corners] (sensor.north) -- ($(sensor.north |- simulink.north) + (0, 3mm)$) -- node[above] {$\mathbf{s}$} ($(simulink.north) + (0, 3mm)$) -- (simulink.north);
\coordinate (anchor-a-x) at ($(simulink.east)!0.5!(sensor.west)$);
\node[] at (anchor-a-y -| anchor-a-x) {$\mathbf{a}$};

%
%
\draw[->, rounded corners] (pi-ppo.east) -- (pi-ppo.east -| simulink-pi.north) -- (simulink-pi.north);

%
%
\tikzset{
  style-boundary/.style={
    dashed
  }
}

\node[fit=(pi-expert-0)(pi-expert-n)(bg-pairs)(sensor), inner ysep=2mm] (top-level) {};
\coordinate (y-top) at (top-level.north);
\coordinate (y-bot) at (top-level.south);
\coordinate (y-top-expert) at (y-top -| bg-trajs.center);
\coordinate (y-top-bc) at (y-top -| pi-bc.center);
\coordinate (y-top-ppo) at (y-top -| pi-ppo.center);
\begin{scope}[on background layer]
\draw[style-boundary] (y-top-expert) -- (y-bot -| bg-trajs.center);
\draw[style-boundary] (y-top-bc) -- (y-bot -| pi-bc.center);
\draw[style-boundary] (y-top-ppo) -- (y-bot -| pi-ppo.center);
\end{scope}
\node[
  text width=33mm, align=center,
  inner sep=0,
] at ($(top-level.north west)!0.5!(y-top-expert) + (0, 2mm)$) {\bf Domain-Randomized Expert Demonstration};
\node[
  text width=60mm, align=center
] at ($(y-top-expert)!0.5!(y-top-bc) + (0, 2mm)$) {\bf BC with \\State-Action Re-Matching};
\node[
  text width=20mm, align=center
] at ($(y-top-bc)!0.5!(y-top-ppo) + (0, 2mm)$) {\bf PPO \\ Fine-Tuning};
\node[
  text width=36mm, align=center
] at ($(y-top-ppo)!0.5!(top-level.north east) + (0, 2mm)$) {\bf Deployment on a Real-World\\Insect-Scale Robot};

\end{tikzpicture}}
    \vspace*{-3mm}
    \caption{Overview of our proposed controller design. First, from a model-based controller, $\pi_{e_i}$, a set of expert demonstrations is generated with randomized domain parameters. Then, we re-match the delayed state with the action to account for system delay. We implement behavior cloning to initialize a neural network controller. Next, in the RL phase, the control policy is fine-tuned with PPO to improve performance and reduce driving command fluctuations. Finally, the controller is integrated into the Matlab Simulink Real-Time environment for demonstrating robot hovering flight.}
    \label{fig:overview}
    \vspace*{-6mm}
\end{figure*}
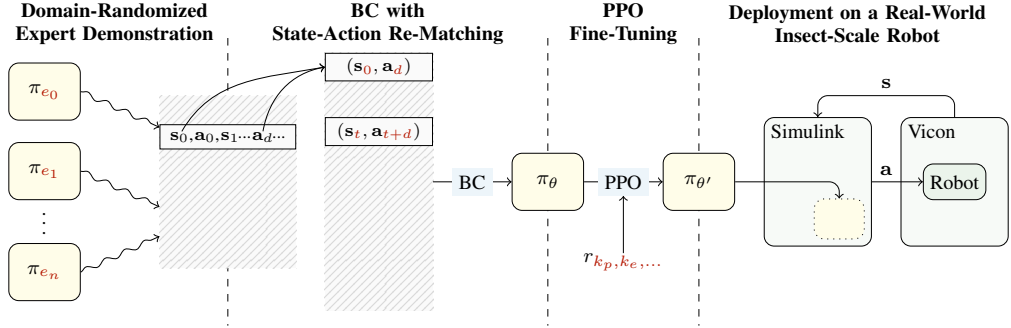

To address these challenges, learning-based methods were applied in several prior studies \cite{perez2015model}\cite{de2021efficient}.  In a previous work \cite{perez2015model}, researchers designed controllers with fixed structures and then used learning methods to identify the control parameters based on flight experiments. Another work \cite{de2021efficient} combined model-based and model-free methods to design a hovering flight controller. The simulations show substantial improvement, but the 1.5-second real-world flight demonstrations suffer a large position error of over 10 cm. This result shows that the unaccounted-for \textit{Sim2Real} gap has a substantial influence on the flight performance of \acp{IMAV} and highlights the difficulties of bringing the controller from simulation to real-world \acp{IMAV}, emphasizing the importance of incorporating model uncertainty and system delay into the simulator.

Another work \cite{tagliabue2023robust} presented a cascaded control architecture, which connects a positional \ac{NN} controller to a model-based attitude controller. 
The researchers used supervised learning to train a \ac{NN} with expert demonstrations from a hand-tuned model predictive controller (MPC). While the flight results show hovering capability, this supervised learning method relies heavily on the performance of the expert controller and may lead to sub-optimal behavior as the time horizon increases \cite{song2023reaching}.  

Here, we propose a deep \ac{RL} approach for controlling \acp{IMAV} and demonstrate stable real-world flights on these soft-actuated platforms. In contrast to previous work \cite{de2021efficient}, we successfully bridge the \textit{Sim2Real} gap and showcase translational flight performance from the simulator to multiple real-world platforms at the insect scale. Compared to \cite{tagliabue2023robust}, our method replaces the entire control scheme with a single \ac{NN} controller and is trained through unsupervised learning (\ac{RL}) to seek optimal performance without relying on hand-tuned expert demonstrations.

Our new approach has two main features: 1) we explicitly account for the system delay of soft IMAVs during the initialization of \ac{NN} by using state-action re-matching in \ac{BC}, and also incorporate this delay in the simulator for \ac{RL} with \ac{PPO}. 2) we randomize domain parameters, including mass, \ac{MoI}, and external disturbances, in both \ac{BC} and \ac{RL} phases to improve controller robustness against system uncertainty. 

These design choices result in a new flight controller that is resilient to system delay and model uncertainty on soft-actuated \ac{IMAV}. We deploy this new type of controller on two distinct DEA-driven IMAVs (Fig. 1) and evaluate their performance. Our results demonstrate multiple zero-shot hovering flights for both robots, marking the first successful deep \ac{RL} flights at the insect scale.  The longest flight we conducted lasts 50 seconds and achieves lateral and altitude \acp{RMSE} of 1.34 cm and 0.05 cm that outperform the state-of-the-art robots of similar scale \cite{bena2023high}\cite{kim2023laser}. By explicitly accounting for system delay and model uncertainty, we achieved substantial flight performance improvement with deep reinforcement learning, representing a significant step toward unlocking the full potential of fast dynamics in soft-driven \acp{IMAV}.


\section{Controller design}

In this section, we describe our flight controller design under the deep \ac{RL} framework. First, we define the robot states and controller actions, together with the dynamics and flight simulator. Next, we develop a modified BC method to initialize the \ac{NN}, which accounts for the system delay and uncertainties by using state-action re-matching and domain-randomized expert demonstrations. Finally, we design a reward function and use \ac{RL} with \ac{PPO} to further optimize the policy and improve driving commands' smoothness. The high-level design of our learning-based controller is illustrated in Fig. \ref{fig:overview}. 


\subsection{States \& Actions} \label{sec:state_action}
The states of our robot include positions $\mathbf{p}=[x,  y, z]^T$, velocities $\mathbf{v}=[\dot{x}, \dot{y}, \dot{z}]^T$, rotational angles represented by quaternions $\mathbf{q}=[q_x, q_y, q_z, q_w]^T$, and angular velocities $\boldsymbol{\omega}=[p, q, r]^T$. The state vector is expressed as: 
\begin{equation*}
    \mathbf{s} = [\; x, \; y, \; z, \; q_x, \; q_y, \; q_z, \; q_w,\; \dot{x}, \; \dot{y}, \; \dot{z}, \; p, \; q, \; r \;]^T,
\end{equation*}
where $\mathbf{p}$, $\mathbf{v}$, and $\mathbf{q}$ are in the world-fixed frame and $\boldsymbol{\omega}$ is on the body frame. The action of the robot is defined as 
\begin{equation*}
    \mathbf{a} = [\;F, \;\tau_x, \;\tau_y\;]^T,
\end{equation*}
where $F$ represents the total thrust force along the body z-axis; $\tau_x$ and $\tau_y$ are the torques with respect to the body x-axis and body y-axis. While our robot cannot generate yaw torque ($\tau_z$), other works \cite{chen2019controlled}\cite{chen2021collision} have shown that hovering flight does not require yaw control authority. 


\subsection{Robot Dynamics \& Simulator}\label{subsec:dynamics_simulator}
The simulator for RL is constructed based on the 6-DOF rigid body dynamics.
Compared to existing UAV simulators, our model accounts for yaw motion damping, external disturbances, and actuation delay.

We aim to develop the \ac{NN} controller in a near-zero yawing condition to simplify training and use the method in \cite{tagliabue2023robust} to re-map actions to the correct body frame. The robot yawing dynamics is thus intentionally constrained by a large damping term $-k_y\Dot{r}$. In addition, the power tethers create external force disturbance ($\mathbf{F}_{dist}$) and torque disturbances ($\tau_{dist,x}$,  $\tau_{dist,y}$) on the robot. The robot dynamics is described by: 
\begin{equation}
    \mathbf{\dot{p}}= \mathbf{v} \label{eq:dy_p2v},
\end{equation}
\begin{equation} 
    \mathbf{\dot{v}} = (\mathbf{R} 
    \begin{bmatrix}
    \;0, \! \! & \! \! 0, \! \! & \! \! F \; 
    \end{bmatrix}^T \!+
    \begin{bmatrix}
    \;0, \! \! & \! \!  0, \! \! & \! \!  \-mg\;
    \end{bmatrix}^T \!+ \mathbf{F}_{dist})/m\label{eq:dy_v2a},
\end{equation}
\begin{equation} \label{eq:dy_ang2angv}
    \mathbf{\dot{q}} = (\mathbf{q}\otimes[ \;0,\;\boldsymbol{\omega}^T\;]^T)/2,
\end{equation}
\begin{equation} \label{eq:dy_angv2angacc}
    \boldsymbol{\dot{\omega}} = 
    \mathbf{J}^{-1}(\-\boldsymbol{\omega}
    \times \! \mathbf{J} 
    \boldsymbol{\omega} + 
    \begin{bmatrix}
        \tau_x \! +\!  \tau_{dist,x}, \! \! \! & \! \!  \tau_y \! +\!  \tau_{dist,y}, \! \! \! & \! \!  \-k_y\dot{r}
    \end{bmatrix}^T),
\end{equation}
where $\mathbf{R}$ is the rotational matrix, $\mathbf{J}$ is the diagonalized moment of inertia tensor, and $\otimes$ is an operator representing quaternion multiplication.

The force generation in the flapping-wing systems, which involves unsteady aerodynamics and the complex underactuated hinge and wing motion, is challenging to model analytically. The inclusion of soft actuator dynamics further complicates this problem. Fortunately, the wing inertia is orders of magnitude smaller than that of the entire robot, allowing the time-averaged lift force to be used for modeling force and torque generation with a slight delay \cite{ma2013controlled}. This delay is typically less than 4 ms, with the overall system delay being dominated by the communication between various instruments required to operate the soft actuators.

To model the overall system delay \cite{tan2018sim2real}, we specify a delay time $d$. The action $\mathbf{a}_t$, which is computed at a time $t$, would be executed on the robot at the time $t+d$; the compact form of robot dynamics can be expressed as 
\begin{equation}\label{eq:delay_dynamics}
    \mathbf{\dot{s}}_{t+d} = f(\mathbf{s}_{t+d}, \, \mathbf{a}_t),
\end{equation}
where the function $f$ represents the nonlinear robot dynamics described in Eq. \eqref{eq:dy_p2v}-\eqref{eq:dy_angv2angacc}.
To solve Eq. \eqref{eq:delay_dynamics} in discrete time, we use the forward Euler method with a step size of 1 ms.


\subsection{Modified Behavior Cloning for Controller Initialization}\label{sec:bc_method}

\begin{figure}[t]
\centering
\centerline{\includegraphics[width=0.485\textwidth]{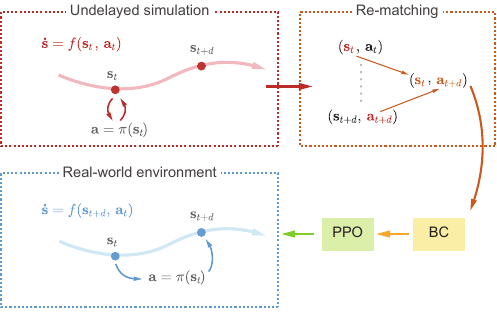}}
\vspace{-2.5mm}
\caption{Workflow of State-Action Re-matching. The expert demonstration is first rolled out in the undelayed simulator; then, we offset the state-action pairs by $d$ and have $(\mathbf{s}_t,\mathbf{a}_{t+d})$ as a pair for supervised learning to clone the delay-compensated controller. The policy then goes through PPO fine-tuning and is deployed to the real-world environment.}
\label{fig:re-match}
\vspace{-6mm}
\end{figure}

To initialize the \ac{NN} controller, we design a modified BC approach that accounts for model uncertainty and system delay. 
We first generate expert demonstrations using a model-based controller \cite{chirarattananon2014single} with randomized robot parameters and disturbances to improve robustness. Then, we re-match the state-action pairs to account for system delay (Fig. \ref{fig:re-match}, upper right). Ultimately, we utilize supervised learning to train an \ac{NN} controller that imitates the expert demonstrations.


\subsubsection{Domain-Randomized Expert Demonstration}
\label{subsec:dr-demo}
To bridge the discrepancy between simulation and real-world environments (\textit{Sim2Real} gap), we randomize three types of domain parameters, including robot parameters $\mathcal{R} = \{ m, I_{xx}, I_{yy} \}$, environmental disturbance parameters $\mathcal{E} = \{ \mathbf{F}_{dist}, \tau_{dist,x}, \tau_{dist,y}$\}, and delay $d$, where $I_{xx}$ and $I_{yy}$ are \ac{MoI} with respect to body x-axis and y-axis.

We specify a value range for each parameter that corresponds to fabrication variation and model uncertainty. For instance, we choose $I_{xx}$ from the interval $[0.75I_{xx}, 1.25I_{xx}]$. Then we pick several $\phi_i$ where each of them is a mapping function from the set $\mathcal{P}$ ($\mathcal{P} \!=\! \mathcal{R} \cup \mathcal{E} \cup \{ d \}$) to their corresponding values.
The closed-loop dynamics under the expert policy $\pi_{e}$ in the environment parameterized by $\phi_i$ becomes
\begin{equation}\label{eq:closedloop_dynamics}
    \mathbf{\dot{s}} = f_{\phi_i}(\mathbf{s}, \, \pi_{e}(\mathbf{s})).
\end{equation}
Based on this domain-randomized closed-loop dynamics, we roll out trajectories with various initial states, $\mathbf{s}_0$, to create expert demonstrations for the training data set.

By incorporating the disturbances (Eq. \ref{eq:dy_p2v}-\ref{eq:dy_angv2angacc}), which push the robot slightly away from the nominal trajectory, we can efficiently sample more states and generate corresponding expert demonstrations during the rollout (Eq. \ref{eq:closedloop_dynamics}) without using iterative \ac{DAgger}\cite{ross2011reduction}.

\begin{figure*}[t]
    \centering
    \centerline{\includegraphics[width=0.995\textwidth]{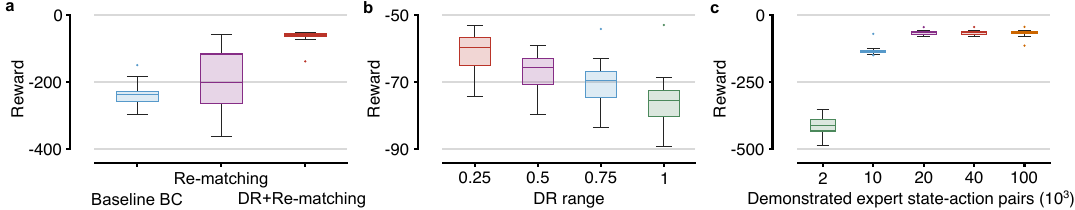}}
    \vspace{-2mm}
    \caption{Simulation results of behavior cloning. (a) Comparison of the baseline method, the method with state-action re-matching, and the method with both state-action re-matching and domain randomization. Colored boxes show 25$\%$, 50$\%$, and 75$\%$ percentiles and the black bars show non-outlier minimum and maximum. Dots are outliers that are 1.5 interquartile range (IQR) away from the top or bottom of the box. (b) Comparison of controller performance as the randomization range increases. (c) Controller performance as a function of training data set size.}
    \label{fig:bc}
    \vspace{-1mm}
\end{figure*}


\begin{figure*}[t]
    \centering
    \centerline{\includegraphics[width=0.995\textwidth]{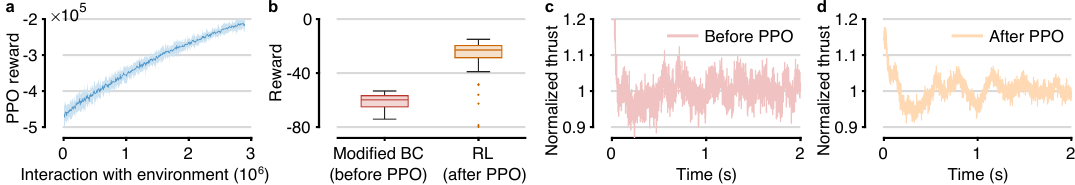}}
    \vspace{-2mm}
    \caption{
    Simulation results of before and after PPO fine-tuning. (a) shows the training curve of the PPO with respect to the chosen reward function. The dark blue line shows the median rewards and the light blue shaded region represents two standard deviations away from the median. (b) displays the performance improvement in simulation after PPO fine-tuning. (c-d) compare the aggressiveness of command before and after PPO, the fluctuation in command is greatly reduced after PPO fine-tuning.
    }
    \label{fig:finetune}
    \vspace{-6mm}
\end{figure*}


\subsubsection{State-Action Re-Matching}\label{subsubsec:rematching}
To account for the system delay, we first turn off delay in the simulator to obtain an undelayed ideal demonstration $\mathcal{T}_{\text{ud}} = \{\mathbf{s}_0, \mathbf{a}_0, ..., \mathbf{s}_t , \mathbf{a}_t , ..., \mathbf{s}_{t+d} , \mathbf{a}_{t+d} ,..., \mathbf{s}_{T} \}$ with a model-based controller (Fig. \ref{fig:re-match}, upper left). 

Next, we "re-match" the state-action pairs.
We recognize that in the real world, an action $\mathbf{a}_t$ would be executed at $\mathbf{s}_{t+d}$ due to system delay (Fig. \ref{fig:re-match}, lower left). Therefore, to have the optimal action $\mathbf{a}_{t+d}$ being executed at $\mathbf{s}_{t+d}$, 
the controller needs to generate $\mathbf{a}_{t+d}$ at $\mathbf{s}_t$. Hence, we re-match the state-action pairs with respect to time and choose $(\mathbf{s}_{t},\mathbf{a}_{t+d})$ as a pair in the training data set $\mathcal{D}$ (Fig. \ref{fig:re-match}, upperleft), where $\mathcal{D}$ is defined as
\begin{equation*}
    \mathcal{D} = 
    \{(\mathbf{s}_{t},\mathbf{a}_{t+d})| \mathbf{s}_t, \mathbf{a}_{t+d}\in \mathcal{T}_{\text{ud}} \text{ where } 0 \le t < T - d \}.
\end{equation*}

\subsubsection{Behavior Cloning (Supervised Learning)}
With the re-matched dataset, $\mathcal{D}$, the \ac{NN} controller $\pi_{\theta}$ can be initialized through \ac{BC} by solving the following optimization equation 
\begin{equation*}
    \max_\theta \sum_{(\mathbf{s},\mathbf{a}) \in \mathcal{D} } \log \pi_\theta(\mathbf{a}|\mathbf{s}).
\end{equation*}
This cloned policy, $\pi_\theta$, has already accounted for the delay from the soft actuators.


\subsection{Reward Function for Reinforcement Learning}\label{sec:reward_func}
To further optimize the \ac{BC} policy, $\pi_\theta$, with \ac{RL}, we design a reward function that considers both states and actions. The state-dependent objective, $r_{s}(\mathbf{s})$, aims to minimize the distance between the current state and the setpoint (origin). To intuitively assign rewards, we use Euler angles ($\phi,\theta,\psi$) retrieved from $\mathbf{q}$. The reward function involves bringing positions ($\mathbf{p}$), velocities ($\mathbf{v}$), two Euler angles ($\phi$ and $\theta$), and two angular velocities ($p$ and $q$) to zero. Since we cannot control the body yaw rate, we do not assign rewards on the states $\psi$ and $r$. The reward function on states is defined as: 
\begin{align*}
r_{s}(\mathbf{s}) =  -(\; &k_p||\mathbf{p}||^2 + k_{e}(||\phi||^2 + ||\theta||^2) + k_v||\mathbf{v}||^2 \\
&+ k_{\omega }(||p||^2 + ||q||^2)\; ),
\end{align*}
where $k_p$, $k_{e}$, $k_v$, and $k_{\omega}$ are hyperparameters that determine the relative weight of each state reward. To specify the action rewards, we penalize aggressive (fluctuating) control outputs and their deviations from the nominal action. The reward functions are defined as 
\begin{align*}
    r_{f}(\mathbf{a}) =  -( \; & k_{ff}||F_t-F_{t-1}||^2 + k_{\tau_xf}||\tau_{x,t}-\tau_{x,t-1}||^2 \\
    &+ k_{\tau_yf}||\tau_{y,t}-\tau_{y,t-1}||^2 \;), \; \forall t \in (0,T],
\end{align*}
\begin{align*}
    r_{n}(\mathbf{a}) = -(k_f||F-F_n||^2 + k_{\tau_x}||\tau_{x}||^2 + k_{\tau_y}||\tau_{y}||^2 ),
\end{align*}
where $k_{ff}$, $k_{\tau_xf}$, $k_{\tau_yf}$, $k_f$, $k_{\tau_x}$, and $k_{\tau_y}$ are hyperparameters that determine the relative weight of each action reward, and $F_n$ is the nominal thrust at the hovering state. The action reward is given by $r_{a}(\mathbf{a}) = r_{f}(\mathbf{a}) + r_{n}(\mathbf{a}).$ The total reward function sums contributions from states and actions:  
\begin{equation*}
    r(\mathbf{s},\mathbf{a}) = r_{s}(\mathbf{s}) + r_{a}(\mathbf{a}).
\end{equation*}


\subsection{Reinforcement Learning through PPO}\label{sec:PPO}
We utilize reinforcement learning to further optimize the policy $\pi_\theta$ that is initialized by modified \ac{BC}. Specifically, we choose to train our policy in the delayed simulator described in Sec. \ref{subsec:dynamics_simulator} (Eq. \ref{eq:delay_dynamics}), whose domain is parameterized by set $\mathcal{P}$ in Sec. \ref{subsec:dr-demo} with various initial conditions, $\mathbf{s}_0$. The \acf{PPO} is implemented to update the policy with the reward function defined in Sec. \ref{sec:reward_func}.

The main challenge in implementing \ac{PPO} involves setting appropriate hyperparameter values in the reward function. Quadrotor-like aerial robots are at least 4th-order systems, where the effects of commanded torques would appear in positional states after being integrated four times. This property makes positional rewards extremely sparse in the policy optimization formulation. As a result, although achieving position control is our primary objective, we still assign rewards to intermediate states such as velocity, Euler angles, and body angular velocity. 

In addition to setting the state rewards, it is also crucial to set appropriate action reward hyperparameters. The actions generated through the \ac{BC} policy are non-smooth \cite{zhao2023learning} because \ac{BC} learns discrete state-action pairs without considering the continuity in time. While fluctuating actions could generate reasonable results in simulation, they are harmful to the lifetime of robotic hardware \cite{hsiao2023heading}. Hence, we assign hyperparameters to the action reward function while minimizing the negative impact on control effectiveness.

The selection of hyperparameter values is guided by an initial educated estimation, followed by empirical tuning. First, we select state-dependent hyperparameters ($k_p$, $k_e$, $k_v$, and $k_\omega$) to ensure each type of state contributes to the reward with comparable magnitude. For instance, position $y = 0.015$ (m) and angular velocity $q = 2$ (rad/s) are considered similarly favorable, so the corresponding $k_p$ and $k_\omega$ are adjusted to ensure they are weighted similarly in the reward function. Action-dependent hyperparameters are chosen in a similar manner. Subsequently, multiple sets of estimated hyperparameters are employed during \ac{PPO} training, and the resulting \ac{NN} policies are evaluated in the simulator. Based on these evaluations, minor adjustments are made iteratively, followed by additional rounds of \ac{PPO} training. 



\subsection{Deployment on Soft-Actuated Robot}
\label{subsec:deployment}
After performing \ac{PPO} fine-tuning, we deploy the \ac{RL}-trained policy, $\pi_{\theta'}$, on our customized experimental setup that runs Matlab Simulink Real-Time at 1 kHz. The flight arena consists of a commercial motion tracking system (Vicon), a specialized controller (Speedgoat), and high-voltage amplifiers (Trek). We built a 720-mg eight-wing IMAV and an 850-mg four-wing IMAV; both of them have four soft actuators (DEAs).


\begin{figure*}
    \centering
    \centerline{\includegraphics[width=0.995\textwidth]{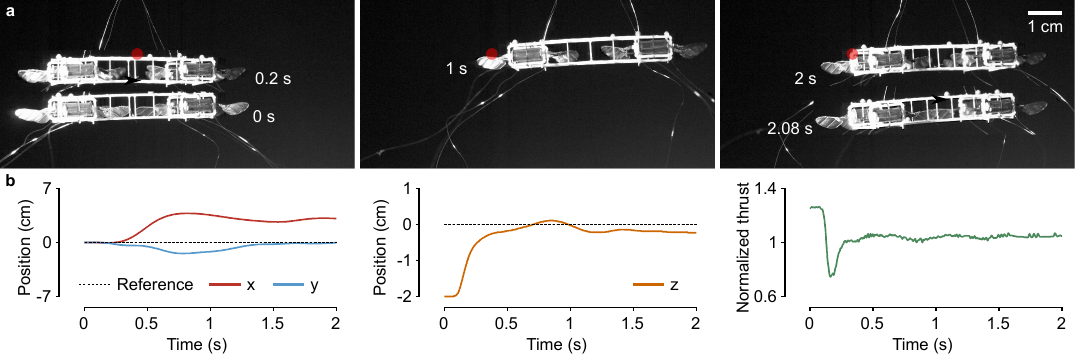}}
    \vspace{-2.5mm}\caption{A successful hovering flight performed by the deep reinforcement learning controller on a 720-mg soft-actuated IMAV. (a) A sequence of composite images illustrating a 2-second hovering flight. (b) Tracked robot lateral position, altitude, and the commanded thrust force.}
    \label{fig:PPO_flight}
    \vspace{-6mm}
\end{figure*}

\section{Results}

We conduct simulations and flight experiments to evaluate controller effectiveness. In simulations, the modified \ac{BC} improves the reward (median) by 75$\%$ compared to the baseline BC. The \ac{PPO} further enhances the reward (median) by 62$\%$ and reduces the thrust fluctuation by 51$\%$, which is crucial for performing experimental validation on real-world hardware. In flight experiments, we achieve multiple zero-shot stable hovering where our position errors outperform the state-of-the-art long endurance flights on IMAVs.


\subsection{Simulation Results}\label{subsec:sim_results}

To train the policy, we use a \ac{NN} with 2 hidden layers, 32 neurons per layer, and $\arctan$ as the activation function. The training and simulations are conducted in the \texttt{Gym} environment on an M1 MacBook Air, with \ac{BC} taking approximately 5 minutes and \ac{PPO} around 15 minutes per iteration. The iterative hyperparameter tuning process for PPO (Sec. \ref{sec:PPO}) is completed in less than a week and results in a set of hyperparameters capable of producing a satisfactory hovering policy.

To compare the \ac{NN} performance in simulation, we run each type of controller in the delayed simulator for 5 seconds and repeat 100 times with different robot parameters $\mathcal{R}$, environmental disturbances $\mathcal{E}$, delays $d$, and initial conditions $\mathbf{s}_0$. 

First, we 
evaluate the effectiveness of the two techniques: state-action re-matching and domain-randomized expert demo. We compare the performance of three behavior cloned controllers: 1) baseline \ac{BC} (non-randomized expert demo and no re-matching); 2) \ac{BC} with state-action re-matching (non-randomized expert demo); and 3) \ac{BC} with both state-action re-matching and domain-randomized expert demonstrations. In Fig. \ref{fig:bc}a, the baseline BC policy returns the lowest median reward of -238.2, while the policy trained on state-action re-matching (without randomized domains) scores -201.2, indicating that the proposed re-matching method improves controller performance in a delayed environment. The policy trained with the randomized domains (with re-matching), achieves the best median reward of -59.8. This result showcases that randomizing parameters at the behavior cloning stage enhances the robustness of the controller to accommodate more model uncertainty.




We also vary the parameter range in our \ac{DR} implementation and evaluate its influence on controller performance. Fig. \ref{fig:bc}b shows that the mean reward remains similar (within 20$\%$ change) despite having large changes in parameter range. This result implies that larger parameter variation can lead to higher tolerance to model uncertainty without substantially sacrificing performance.

Furthermore, we investigate the learning convergence rate. Fig. \ref{fig:bc}c shows the controller performance improves as the number of training state-action pairs increases; specifically, the median rewards for 2,000, 10,000, and 20,000 state-action pairs are -409.8, -134.9, and -66.1, respectively. However, further increasing the number of training pairs has a diminishing effect on the rewards, as 40,000 and 100,000 pairs yield rewards of -65.9 and -64.7, respectively, an improvement of only 0.2 for additional 20,000 data points. This result shows training the policy only requires approximately 20,000 state-action pairs to converge, equivalent to 20 seconds of expert flight demonstrations in the simulation. 



In addition, we investigate the performance of PPO fine-tuning and its influence on smoothing the control policy at the \ac{RL} stage. Fig. \ref{fig:finetune}a shows the mean reward increases as the number of agent-environment interactions increases. Fig. \ref{fig:finetune}b demonstrates that the fine-tuned  \ac{RL} policy, $\pi_{\theta'}$, is improved through \ac{PPO} by 62$\%$ (median reward). Fig. \ref{fig:finetune}c and \ref{fig:finetune}d compare the commanded thrust before and after PPO fine-tuning. The command fluctuation reduces by 51$\%$ without reducing trajectory tracking accuracy, making it more preferable to be deployed on real-world hardware.


\subsection{Experimental Flight Results}\label{sec:result_flight}


\begin{figure*}
    \centering
    \centerline{\includegraphics[width=0.995\textwidth]{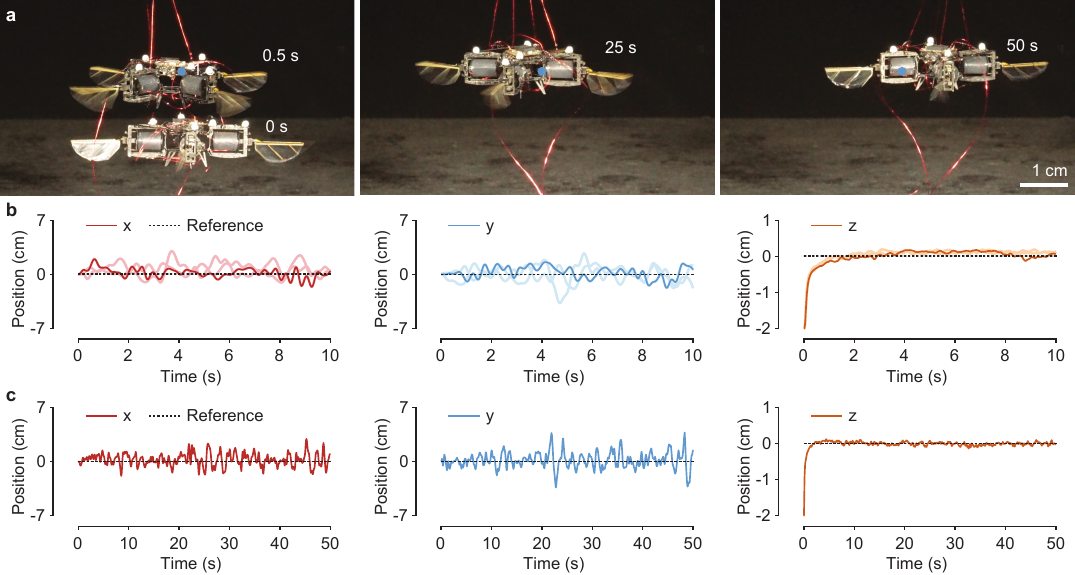}}
    \vspace*{-2.5mm}\caption{Successful hovering flights performed by the deep reinforcement learning controller on an 850-mg soft-actuated four-wing IMAV. (a) A sequence of composite images illustrating a 50-second hovering flight. The blue dots in the images indicate the setpoint (origin) of the robot. (b)-(c) Tracked robot lateral position and altitude. (b) Three 10-second hovering flights. The light colors represent repeating flights. (c) The 50-second hovering flight.}
    \label{fig:new_flight}
    \vspace{-6mm}
\end{figure*}

To evaluate the effectiveness of state-action re-matching and \ac{DR} in bridging the \textit{Sim2Real} gap, we deploy the trained policy on two distinct soft-actuated robots. The results demonstrate successful and stable hovering flights on both real-world platforms.

We first conduct a flight test on a 720-mg eight-wing \ac{IMAV} (Fig. \ref{fig:robot_photo}, left) to evaluate the reinforcement learning policy after PPO fine-tuning (Fig. \ref{fig:PPO_flight}). The soft-actuated robot achieves a zero-shot stable hovering with small lateral drift, marking the first successful deployment of deep \ac{RL} control on an insect-scale flapping-wing soft robot (Supplementary Video - Flight Video 1). Fig. \ref{fig:PPO_flight} illustrates the tracked position and commanded thrust, with position errors comparable to other work \cite{chen2019controlled}\cite{chen2021collision}. The commanded thrust is also reasonably smooth, highlighting the effectiveness of PPO fine-tuning.

We then attempt to fly a four-wing robot (Fig. \ref{fig:robot_photo} right), which has an \ac{MoI} approximately six times smaller than the eight-wing version on the y-axis, making it even more challenging to control. To adapt to this design, we adjust only the robot parameters $\mathcal{R}$ for both \ac{BC} and \ac{PPO}, keeping other parameters unchanged for \ac{NN} controller training. Using the trained policy, we achieve three consecutive 10-second flights (Fig. \ref{fig:new_flight}b and Supplementary Video - Flight Video 2) with lateral position and altitude \acp{RMSE} of 0.97–1.58 cm and 0.10–0.12 cm, respectively (error calculated after a 1-second takeoff stage to allow altitude to converge). To further evaluate the policy’s reliability, we conduct an extended 50-second flight—longer than any other reported flight at the insect scale \cite{hsiao2023heading}. During this flight, the lateral position and altitude \acp{RMSE} are 1.34 cm and 0.05 cm, respectively (Fig. \ref{fig:new_flight}a,c and Supplementary Video - Flight Video 3, error calculated after a 1-second takeoff).

Compared with other long hovering flights ($>$ 10 s), the position errors of these successful flight attempts on the four-wing soft-actuated robot are smaller than those reported in state-of-the-art  \ac{IMAV} studies \cite{ bena2023high}\cite{ chen2021collision}\cite{kim2023laser}.

\section{Discussion \& Conclusion}

In this work, we develop a deep reinforcement learning-based controller for insect-scale aerial robots. We address the challenges of system delay and model uncertainty by initializing the policy through the state-action re-matching method with domain-randomized demonstrations. Then, we apply PPO in the reinforcement learning stage to improve flight performance and reduce driving command fluctuation. The simulation results show that the proposed techniques for BC can effectively improve the mean reward, and PPO fine-tuning reduces variations of thrust. Most importantly, we deploy this controller on a 720-mg and an 850-mg soft-actuated IMAV and demonstrate a 50-second hovering flight with lateral position and altitude error of 1.34 cm and 0.05 cm, respectively.

Achieving insect-like locomotion on a soft-actuated \ac{IMAV} requires complex planning and high-rate feedback control, yet we are limited to lightweight onboard microprocessors with constrained computational capacity. Deep reinforcement learning is an ideal solution, as a small (32x32) multi-layer perceptron can run efficiently on hardware of this size, and the neural network can learn an optimal policy over long time horizons through deep \ac{RL} \cite{song2023reaching}.

Unlike most \ac{BC} methods that rely on \ac{DAgger} \cite{ross2011reduction} to enhance neural network robustness after cloning, the proposed modified \ac{BC} method incorporates disturbances, $\mathcal{E}$, directly during the expert demonstration stage. This early introduction allows us to sample more diverse state-action pairs without \ac{DAgger}, making demonstration generation computationally efficient and reducing dataset creation time to under a minute. Including supervised learning, the modified BC approach initializes a delay-compensated \ac{NN} ready for PPO in less than 5 minutes (on an M1 MacBook Air).

Compared to \cite{tagliabue2023robust}, the proposed modified \ac{BC} could generate demonstrations more efficiently for long trajectories. For dynamically feasible trajectories \cite{sun2022comparative}, model predictive controls (MPC) \cite{tagliabue2023robust} is often computationally intensive and time-consuming. In contrast, the proposed modified \ac{BC} method, which involves no optimization, could efficiently initialize a \ac{NN} and leave the long-horizon optimization to the \ac{RL} stage \cite{song2023reaching}. At the same time, the policy obtained through deep \ac{RL} would not be limited by the sub-optimal demonstration from the hand-tuned MPC.

An essential aspect of this work is bridging the \textit{Sim2Real} gap. For the first time, an insect-scale robot achieves stable hovering flight using model-free deep \ac{RL}, demonstrating that controllers proven effective in the simulator can reliably translate to real-world soft-actuated robots. This milestone not only validates the robustness of our approach but also paves the way for testing more control strategies safely and efficiently within the simulation environment.

While this work focuses on hovering flights, it represents an intermediate step toward achieving insect-like agile maneuverability. By harnessing the potential of unsupervised deep reinforcement learning, the neural network controller can be trained on more complex tasks in simulation—such as wall perching \cite{chirarattananon2016perching}, inverted ceiling landings \cite{habas2023inverted}, and aggressive trajectory following \cite{song2023reaching}—and perform these challenging maneuvers on real-world IMAVs in the near future.



\vspace*{1mm}


\balance
\bibliographystyle{IEEEtran}

\end{document}